\pdfoutput=1
\documentclass[10pt,twocolumn,letterpaper]{article}

\usepackage{cvpr}
\usepackage{times}
\usepackage{epsfig}
\usepackage[breaklinks=true,bookmarks=false]{hyperref}
\usepackage{capt-of}
\usepackage{color}
\usepackage{algpseudocode}
\usepackage{algorithm}

\usepackage{graphics, graphicx}
\usepackage{subfigure}
\usepackage{epstopdf}
\usepackage{float}
\usepackage{multirow}
\usepackage{cite}
\usepackage{latexsym,bm,amsmath,amssymb}
\usepackage{enumerate}
\usepackage[noblocks]{authblk}
\usepackage{flushend}

\cvprfinalcopy 

\begin{document}

\title{Who Leads the Clothing Fashion: Style, Color, or Texture? \\ A Computational Study}

\author[1]{Qin Zou}
\author[1]{Zheng Zhang}
\author[1]{Qian Wang\thanks{Corresponding author.}}
\author[2]{Qingquan Li}
\author[3]{Long Chen}
\author[4]{Song Wang}
\affil[1]{\small School of Computer Science, Wuhan University, P.R.~China}
\affil[2]{Shenzhen Key Laboratory of Spatial Smart Sensing and Service, Shenzhen University, P.R.~China}
\affil[3]{School of Data of Computer Science, Sun Yat-sen University, P.R.~China}
\affil[4]{Department of Computer Science and Engineering, University of South Carolina, USA}
\maketitle
\begin{abstract}
It is well known that clothing fashion is a distinctive and often habitual trend in the style in which a person dresses. Clothing fashions are usually expressed with visual stimuli such as style, color, and texture. However, it is not clear which visual stimulus places higher/lower influence on the updating of clothing fashion. In this study, computer vision and machine learning techniques are employed to analyze the influence of different visual stimuli on clothing-fashion updates. Specifically, a classification-based model is proposed to quantify the influence of different visual stimuli, in which each visual stimulus's influence is quantified by its corresponding accuracy in fashion classification. Experimental results demonstrate that, on clothing-fashion updates, the style holds a higher influence than the color, and the color holds a higher influence than the texture.
\end{abstract}

\section{Introduction}
Fashion is a product of social psychology which reflects the common aesthetics that people share with each other. Clothing is probably one of the most salient and expressive mediums to construct and deliver fashion concepts. The clothing-fashion updates can provide useful sources to understand the aesthetics and its evolution in the human society, which holds great importance in scientific research. Meanwhile, accurate predictions of clothing-fashion updates would benefit the clothing industries in guiding them producing fashionable clothes and avoiding wasting.

From the perspective of visual perception, clothing can be defined by its style, color and surface texture, which are three different kinds of visual stimuli to human visions. It complies with the findings in previous study that objects in the natural world possess different visual attributes, including shape, color, surface texture and motion~\cite{bloj1999na}. Many people know these visual stimuli could place some influence on the updating of clothing fashion, but few can clearly tell or explain which one places more influence than the other. With the development of computer vision and machine learning technologies, it is possible to parse visual stimuli of clothing into machine-understandable patterns, e.g., shape patterns, color patterns and texture patterns, etc. These visual patterns can be used by a learning machine for training, testing and classification, based on which we can quantize the influence each visual stimulus places on the updating of clothing fashion.

Style is one distinctive visual stimulus that influences the clothing-fashion updates. As we know, in a broad sense the clothing color and texture also contribute to the clothing style. In this work, the clothing style is defined in a narrower sense, and it refers to the structure and shape of the clothing. Shape features of objects can often provide robust and efficient information to make them identifiable~\cite{iqbal2012content}. It can be easily observed that, the long coat, T-shirt, vest and skirt are very different clothing styles. Even for one specific theme, e.g., the skirt, there may exist a wide variety of styles: winding skirt, bell-shaped skirt, flared skirt, fishtail skirt, cheongsam skirt, muffled skirt, and bubble skirt, etc. While clothing styles evolve in human dressing history, the fashion styles vary from time to time. For example, the cheongsam dresses were very popular in Hong Kong in the 1950s, while the tunic suits were in great fashion in Beijing in the 1980s. The influence of style on clothing fashion can also be found in the fashion shows, where a number of new clothing styles would be released to the public each year, which are then spread around the world.

Color is another visual stimulus that influences the fashion trend of clothing. The color of objects or scenes does play an important
role in vision~\cite{wu2014color}. Color is considered as one of the most expressive visual features, which can often make objects more salient and recognizable~\cite{hayakawa2011snake}.
Beside, color can influence people's mental, emotional, and feelings, e.g., the red means warm, hot, and passion, the blue means cold and indifferent, and the grass green means hope and growth, etc. Color can play as a special language for clothing during artistic and aesthetic experiences~\cite{spence2010color,khoroshikh2011color,birren1950color}, and is an important feature of clothing and clothing fashion. The changing of colors could commonly be found when clothing fashion updates. For example, the worm red was fashioned in 2010, while the military green was very popular in 2011.

Texture, i.e., the surface texture which shows regular details in an image~\cite{fernandez2016texture}, is the third visual stimulus that can affect the clothing-fashion updates. Generally, different textures can be produced by using different materials, or by varying the manufacture processes or technologies. These changing textures would result in different visual stimuli to human visions, and bring different aesthetic feelings. Thus, in clothing-fashion updates, the texture can also be an important influential factor. For example, the chiffon was very popular in 2010, while the taffeta was in great favor in 2013.

The purpose of this study is to discover the importance of the above three visual stimuli, i.e., style, color and texture, with respect to their power of influence on clothing-fashion updates. In this work, we use fashion classification to indirectly assess the importance of these three visual stimuli to fashion updates. Clothing fashions usually update season by season. In the paper, we study the use of each visual stimulus as features to recognize the underlying fashion season of clothes. The more accurate the recognition, the more influential this visual stimulus in clothing fashions. More specifically, we collect 9,339 representative photos of fashioned clothes in 8 continuous years (2008-2015)  from 14 world-famous brands and each year is treated as a different fashion season. This way, these photos are categorized into 8 classes based on their season labels. We employ the classical techniques in computer vision and machine learning to perform this 8-class fashion classification and use it to evaluate the importance of each visual stimulus.
For each kind of visual stimuli, the according features are extracted and fed into a learning machine for training and testing. Then, the visual stimulus that produces more effective features for classification will be considered as the one holds higher influence power on clothing-fashion updates. Specifically, shape features based on the Scale Invariant Feature Transform (SIFT) descriptors~\cite{lowe2004sift} are used to encode the clothing styles, color features based on the Color Name (CN) descriptors~\cite{Benavente2008cn} are used to encode the clothing colors, and texture features based on the Local Binary Patterns (LBP) descriptors~\cite{ojala2002lbp} are used to encode the clothing textures.

\begin{table*}[!htbp]
\caption{FASHION8 contains 9,339 photographs of 14 world-famous brands in the recent 8 years.}\label{tbl:fashion8}
\vspace{0.1in}
\scriptsize
\centering
\begin{tabular}{|c|c|c|c|c|c|c|c|c|c|c|c|c|c|c|c|c|}
\hline
  \multirow{4}{0.5cm}{Year} & \multirow{4}{0.80cm}{Bottega Veneta} & \multirow{4}{0.9cm}{Burberry Prorsum} & \multirow{4}{0.75cm}{Dior Homme} & \multirow{4}{0.9cm}{Emporio Armani} & \multirow{4}{0.8cm}{Erme. Zegna} & \multirow{4}{0.9cm}{Givenchy} & \multirow{4}{0.55cm}{Gucci} & \multirow{4}{0.7cm}{Hermes} & \multirow{4}{0.6cm}{John Varvat} & \multirow{4}{0.55cm}{Louis Vuitto} & \multirow{4}{0.6cm}{Neil Barret} & \multirow{4}{0.5cm}{Prada} & \multirow{4}{0.55cm}{Thom Brow} & \multirow{4}{0.75cm}{Versace} & \multirow{4}{0.6cm}{Sum}\\
  & & & & & & & & & & & & & & & \\
  & & & & & & & & & & & & & & & \\
  & & & & & & & & & & & & & & & \\
  \hline
  2008 & 78 & 80 & 51 & 187 & 0 & 38 & 87 & 87 & 46 & 109 & 70 & 83 & 103 & 80 & 1099 \\
  \hline
  2009 & 54 & 87 & 79 & 202 & 58 & 57 & 89 & 46 & 36 & 94 & 70 & 89 & 111 & 92 & 1164 \\
  \hline
  2010 & 79 & 78 & 80 & 155 & 62 & 55 & 89 & 80 & 49 & 104 & 67 & 74 & 81 & 86 & 1139 \\
  \hline
  2011 & 74 & 83 & 75 & 157 & 103 & 70 & 85 & 43 & 81 & 98 & 71 & 79 & 83 & 92 & 1194 \\
  \hline
  2012 & 74 & 79 & 79 & 148 & 81 & 79 & 73 & 83 & 70 & 79 & 73 & 84 & 86 & 91 & 1179 \\
  \hline
  2013 & 86 & 85 & 79 & 135 & 73 & 57 & 79 & 66 & 65 & 79 & 68 & 76 & 79 & 101 & 1128 \\
  \hline
  2014 & 88 & 85 & 82 & 192 & 79 & 100 & 75 & 81 & 71 & 83 & 76 & 73 & 82 & 82 & 1249 \\
  \hline
  2015 & 82 & 85 & 80 & 150 & 80 & 104 & 72 & 72 & 75 & 76 & 76 & 65 & 81 & 89 & 1187 \\
  \hline
\end{tabular}
\end{table*}

\begin{figure*}[!ht]
\centering
\hspace{-0.2in}
\includegraphics[width=0.8\linewidth]{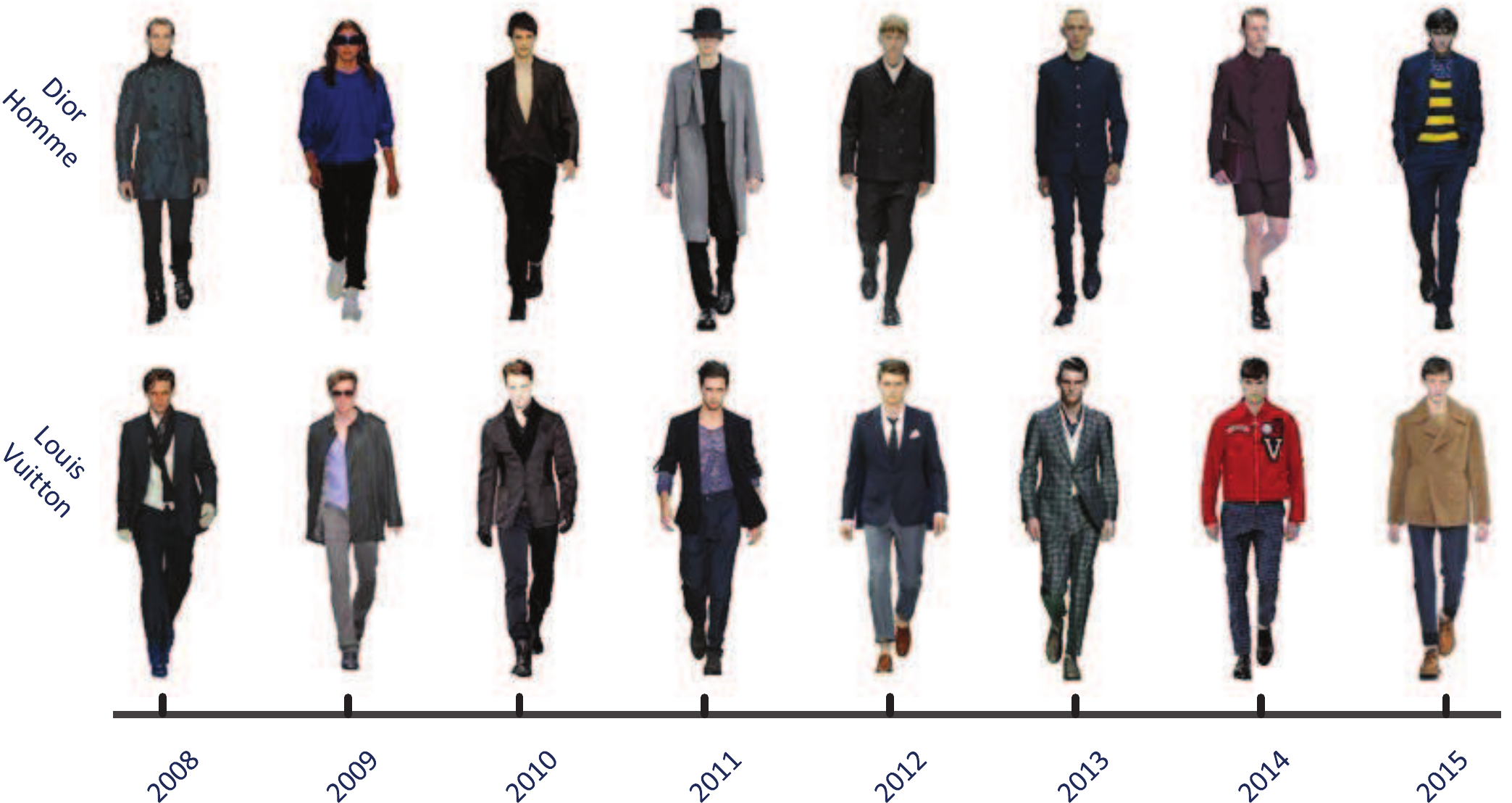}
\caption{\textbf{Example photographs of two brands in FASHION8: Dior Homme and Louis Vuiton, from 2008 to 2015.}}
\label{fig:fashion8}
\end{figure*}

\section{Methods}

\noindent
\textbf{Data and Preprocessing.} There are four datasets used in our experiments, i.e., STYLE130, COLOR335, Outex-TC-00021 and FASHION8. Detailed inform of these datasets has been given below.

\begin{figure*}[!htbp]
\centering
    \subfigure[]{
        \label{style:image}
        \includegraphics[width=0.38\linewidth]{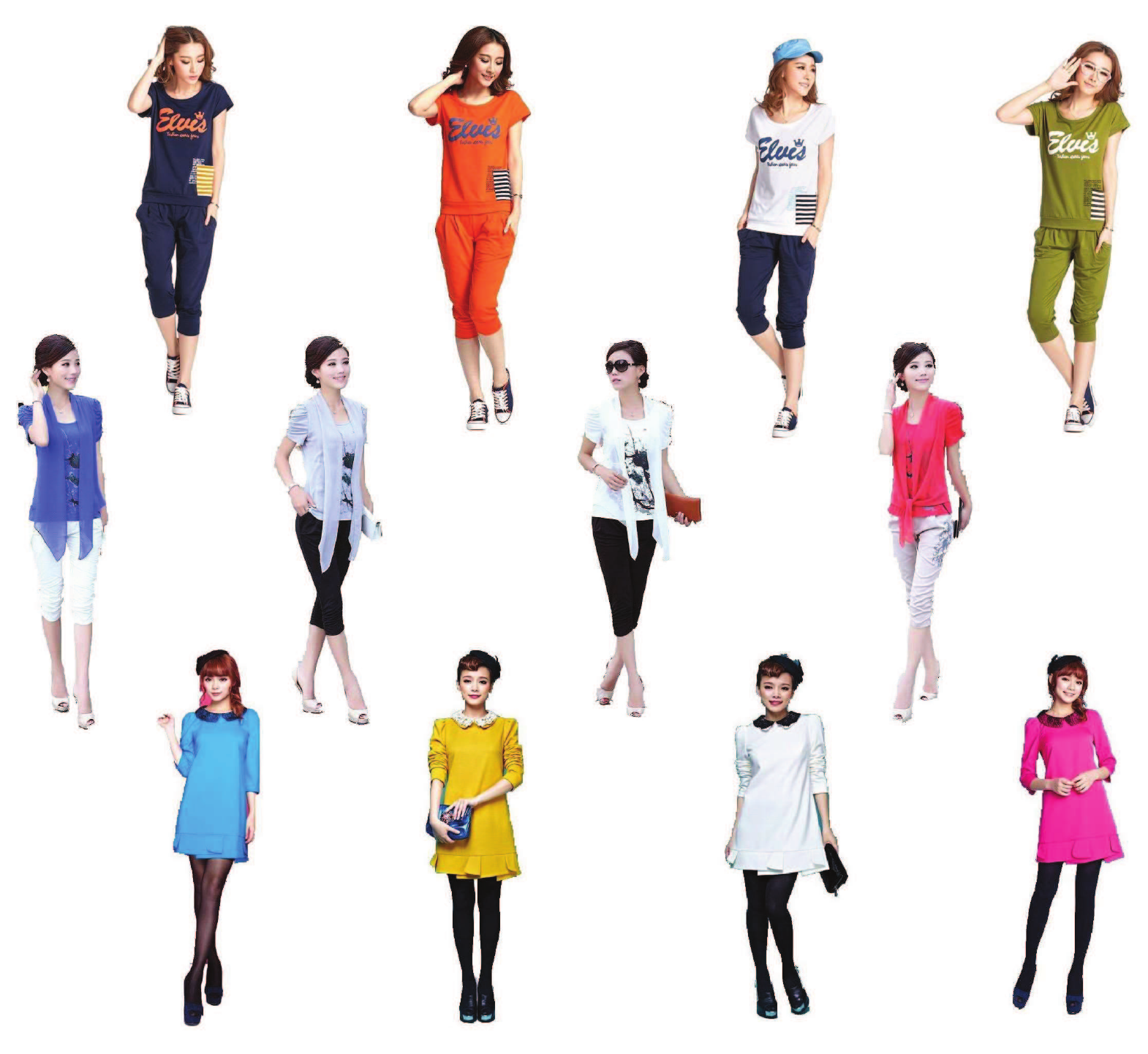}}
    \hspace{0.28in}
    \subfigure[]{
        \label{style:precision}
        \includegraphics[width=0.46\linewidth]{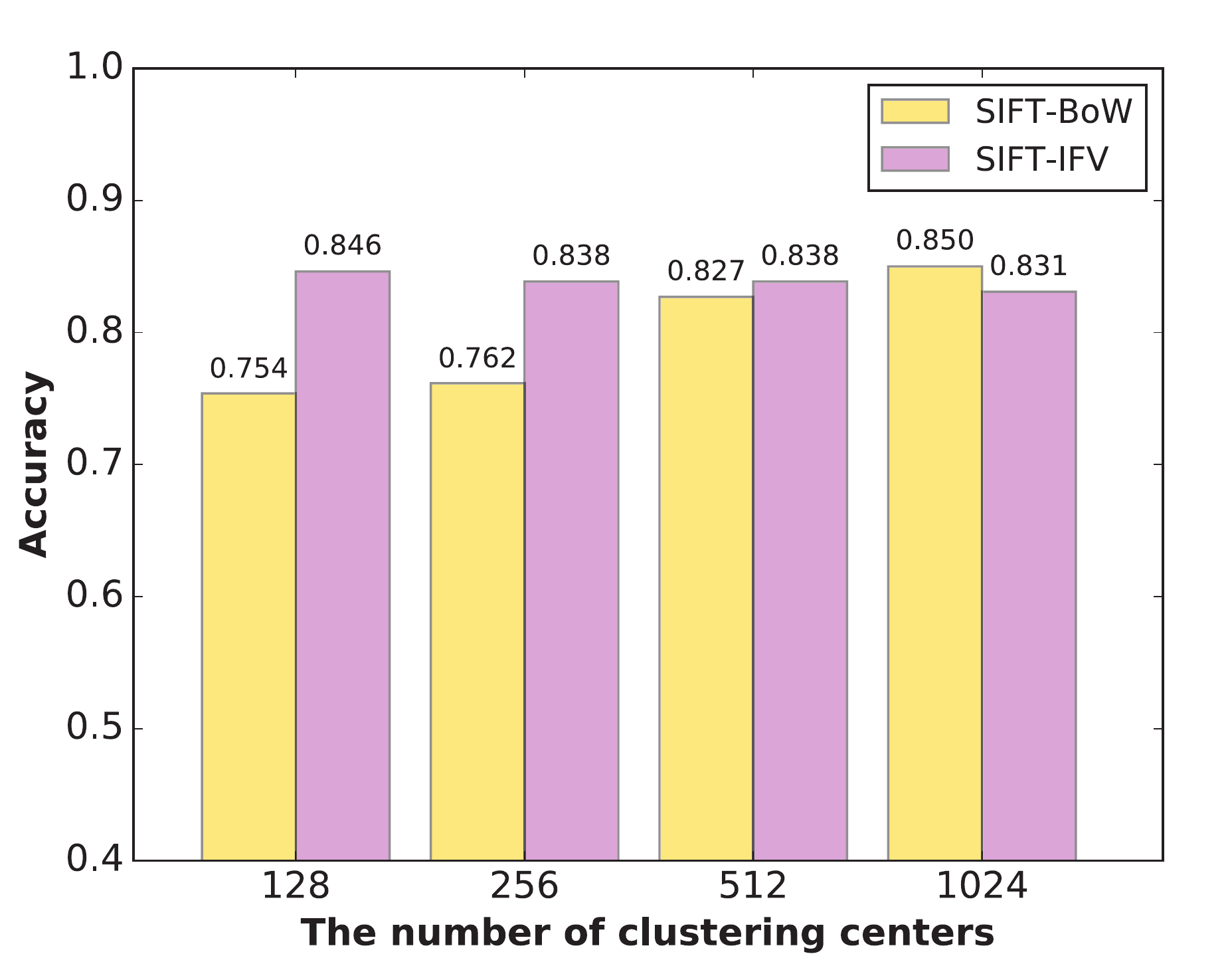}}
    \caption{\textbf{Performance on clothing style classification.}
    (a) Three sample categories in dataset STYLE130, with each row one category.
    (b) Performances of clothing style classification using `SIFT-BoW' and `SIFT-IFV', running with $numC$=128, 256, 512, and 1024, respectively.
    }
\end{figure*}

\begin{itemize}
\item \textbf{STYLE130} - it contains 130 different clothing styles. One clothing style in STYLE130 corresponds to one category, which contains four clothing photographs with similar clothing structures or shapes as labelled by human. For each category, two images are used for training and the remaining two for testing. Example categories for three clothing styles are shown in Fig.~\ref{style:image}.

\item \textbf{COLOR335} - it contains 335 different clothing color categories. Each category in COLOR335 includes three clothing photographs with similar color collocation. Examples of four categories are shown in Fig.~\ref{color:image}. For each category, two images are used for training and the remaining one for testing.

\item \textbf{Outex-TC-00021} - it is a publicly available texture dataset, which contains 2,720 images of 68 different texture types, with 40 images per type. Sample images of nine texture types are shown in Fig.~\ref{texture:image}. One half of the images are used for training and the remaining half are used for testing.

\item \textbf{FASHION8{\footnote{The dataset is available at https://sites.google.com/site/qinzoucn}}} - it contains 9,339 fashion photographs, which are collected from fashion shows of 14 world-famous brands in the recent 8 years (2008 -- 2015). Sample images of two brands are shown in Fig.~\ref{fig:fashion8}. The names of the 14 brands and the number of photographs each brand provides have been listed in Table~\ref{tbl:fashion8}. Fashion photographs from the same year are defined as one category.
\end{itemize}

To avoid the effect of scale, we resize each image to contain the human body with nearly the same height in the image, and then we crop the image region to just contain the human body. In Exp.~1 and Exp.~2, the photographs are resized and cropped to the same height of 800 pixels. In Exp.~3, texture images in Outex-TC-00021 have the same size of 128$\times$128 pixels. In Exp.~4, all photographs in FASHION8 share the same size of 381$\times$768 pixels and just contain the human body. To avoid the effect of background contexts in classification, we create a mask for each image to annotate the foreground human. We develop a human-computer interaction tool to segment the foreground in the image, using the Graph-cut-based algorithm~\cite{boykov2001iccv}. Finally, STYLE130 contains 130 different clothing styles with four clothing photographs of same style in each style. COLOR335 includes 335 categories and in each category, there are three clothing photographs with similar color collocations. Outex-TC-00021 consists of 68 different texture types with 40 images per type, and FASHION8 contains 9,339 fashion photographs in recent eight years, from 2008 to 2015, with each year defined as one category.

\noindent
\textbf{Procedure.}
In order to find the appropriate descriptors and validate their capabilities in describing the corresponding visual stimuli, we first conduct Exp.~1, Exp.~2, and Exp.~3 for the style-, color- and texture- based clothing classification, respectively. Then we conduct Exp.~4 on the FASHION8 dataset, by selecting the top-three or four best performed descriptors in the above three experiments to do the fashion classification.

After removing the image background, we extract the image features for feature encoding and image representation. For style descriptor - SIFT, a dense sampling strategy is employed. Specifically, multi-scale dense SIFTs are extracted from each input image, with a sampling step of 4, and a scale number of 5. For color descriptors - CN and RGB, features are extracted with a sampling step of 5, and a scale number of 2. Note that, CN produces 11-dimensional features.  For texture descriptors, eight types of LBPs are computed at each pixel in the image. Once a set of local features are extracted from a region, a feature encoding procedure is applied on them to build the final feature vectors. For SIFT, we follow the standard BoW and IFV feature-encoding algorithms. For CN and RGB features, a regional color co-occurrence framework (RCC~\cite{zou2015icip}) is employed. For features from each type of LBP, we directly project them into a histogram with their pre-defined pattern tables~\cite{ojala2002lbp}. Before building the feature histogram, in BoW, a feature codebook is constructed for each type of features by $k$-means clustering, and followed by a hard-assignment strategy for feature quantization. While in IFV, features are supposed to be represented by a number of $n$ Gaussian models, which can be treated as $n$ clustering centers. In Exp.~1, a number of 128, 256, 512 and 1024 clustering centers are tested for both Bow-SIFT and IFV-SIFT. In Exp.~2, a number of 64, 128 and 256 clustering centers are tested for CN, CN-RCC, RGB-RCC. In Exp.~1, two images of each category are used for training and the remaining two images are used for testing. In Exp.~2, two images of each category are used for training and the remaining one is used for testing. In Exp.~3, twenty images of each category are randomly selected and used for training, and the remaining twenty images are used for testing. In Exp.~4, the training data are selected randomly with consistent the same ratio in each category and we repeat this process thirty times to get statistically reliable results. For a fair comparison, we vary the amount of training data from 10\% to 90\% at an interval of 10\%.

We employ a multi-class support vector machine (SVM~\cite{burges1998svm,guyon2002ml}) for classification, using a one-vs-all strategy. As each sample holds a label, we could evaluate the performance of feature descriptor by comparing the predicted labels with the ground-truth labels. The accuracy (\emph{Acc}) is defined as $\emph{Acc} = \frac{TP}{TT}$, where $TP$ is the number of the correctly predicted samples, and $TT$ is the total number of the testing samples.

\section{Results}
\subsection{The capability of shape descriptors in clothing-style description}
 In Exp.~1, we investigate the capability of the shape descriptor - SIFT (Scale Invariant Feature Transform~\cite{lowe2004sift}) in describing the clothing styles. The dataset - STYLE130 - is used for evaluation. Examples of three categories are shown in Fig.~\ref{style:image}. For each category, two images are used for training and the remaining two for testing. For classification, multi-scale dense SIFT features are extracted from each photograph, and two famous feature-encoding algorithms are employed for image representation, namely the Bag-of-Words~\cite{sivic2003bow} (BoW) and the Improved Fisher Vector~\cite{perronnin2007fisher,perronnin2010eccv} (IFV). For convenience, the resulting classification methods are then named as `SIFT-BoW' and `SIFT-IFV', respectively. Both methods are tested by using 128, 256, 512, and 1024 clustering centers, respectively. The average accuracies are calculated on the classification results of all 130 styles. The results are plotted in Fig.~\ref{style:precision}, from which we can see that, for `SIFT-BoW', the classification accuracy increases monotonically with the number of clustering centers ($numC$), and it reaches 85.0\% at $numC$=1024. For `SIFT-IFV', the classification accuracy is stably above 83.0\% when we vary the number of clustering centers, and it reaches the highest of 84.6\% at $numC$=128. Therefore, `SIFT-BoW' and `SIFT-IFV' obtain an accuracy about 85\% when classifying a total of 130 different clothing styles, which demonstrates their high capability in clothing-style description.

\begin{figure*}[!htbp]
\centering
    \subfigure[]{
        \label{color:image}
        \includegraphics[width=0.38\linewidth]{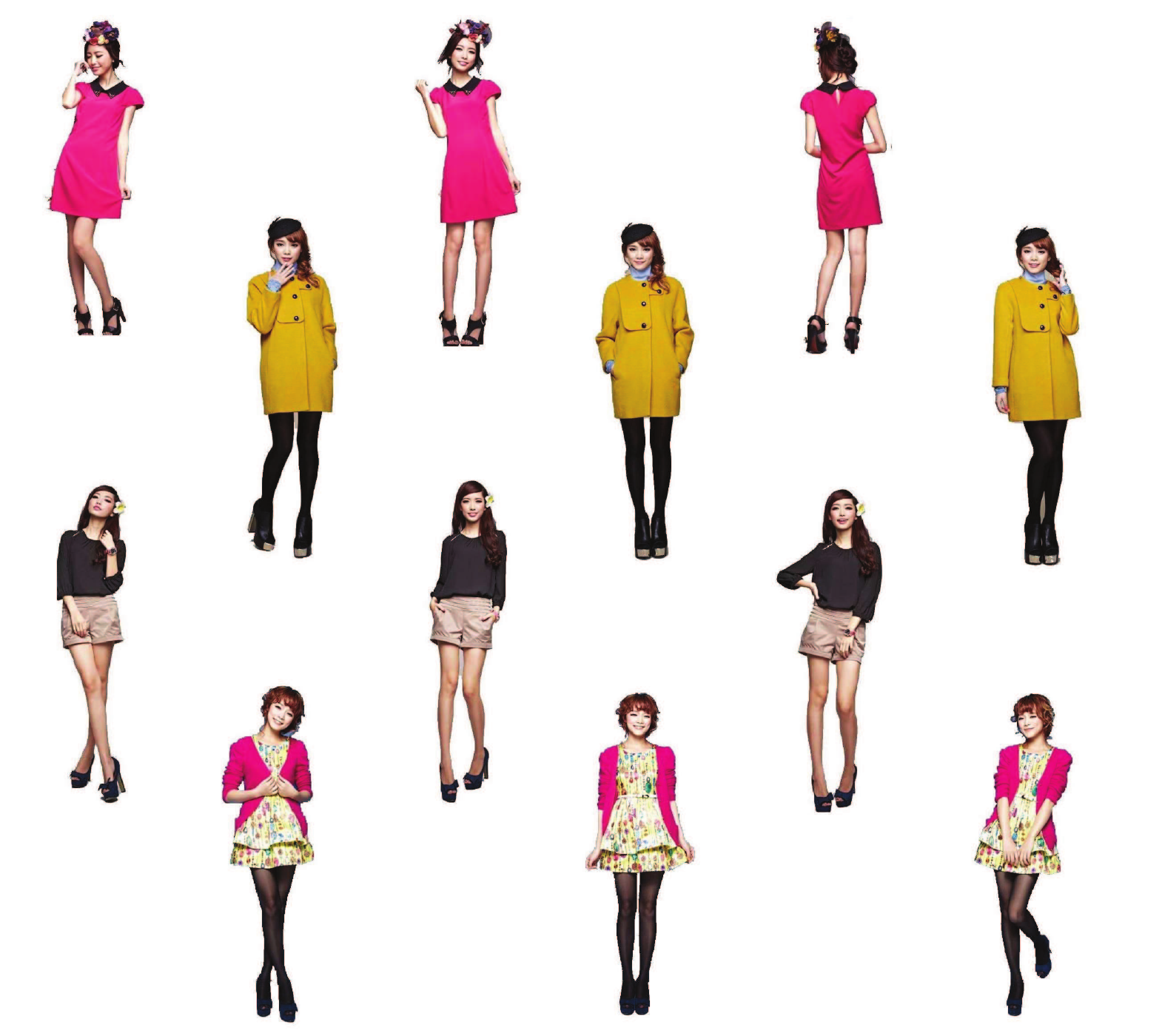}}
    \hspace{0.28in}
    \subfigure[]{
        \label{color:precision}
        \includegraphics[width=0.46\linewidth]{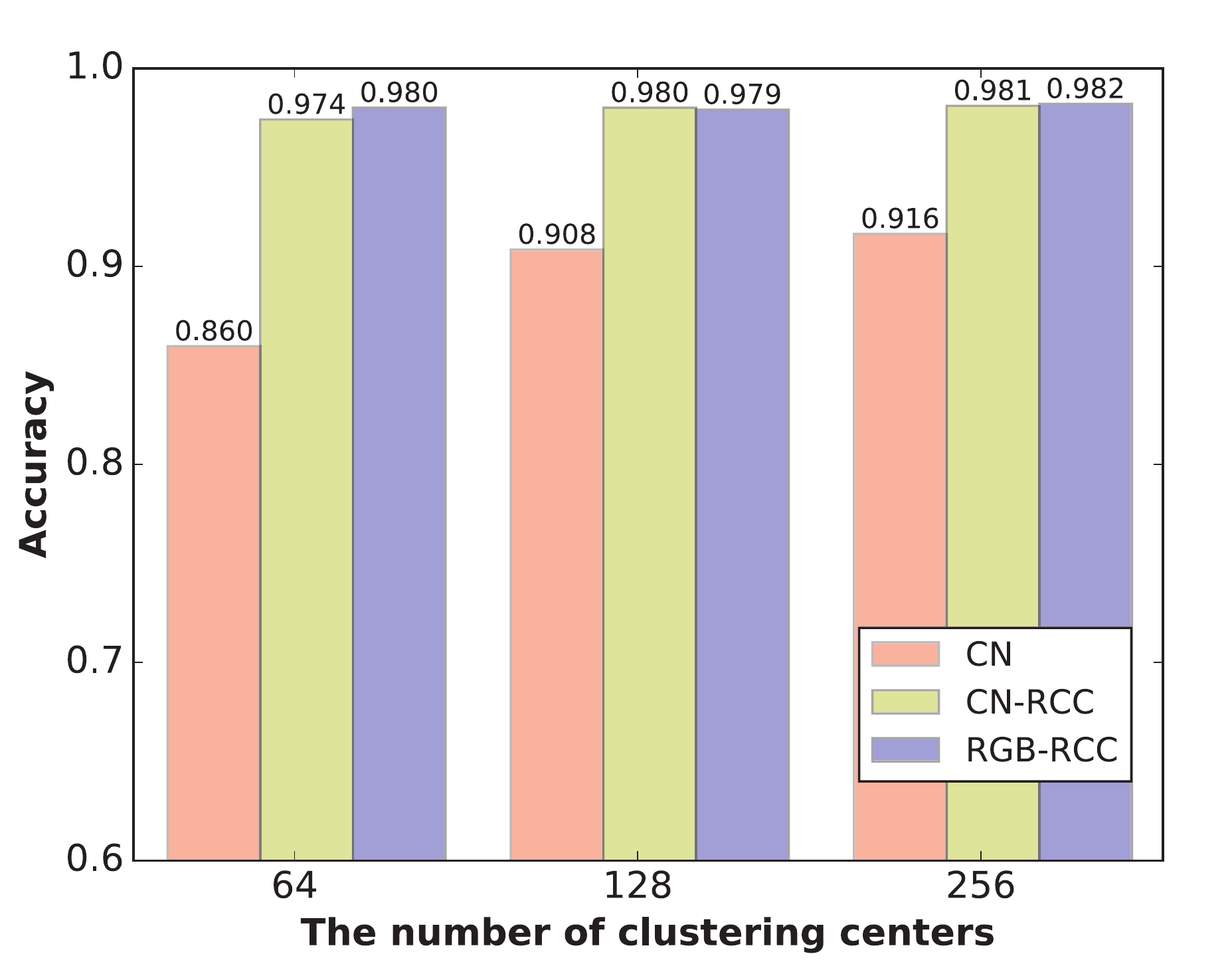}}
    \caption{\textbf{Performance on clothing color classification.}
    (a) Four sample categories in dataset COLOR335, with each row one category.
    (b) Performances of clothing color classification using `CN', `CN-RCC' and `RGB-RCC', running with 64, 128, 256 clustering centers.
    }

\end{figure*}

\subsection{The capability of color descriptors in clothing-color description}
 In Exp.~2, we examine the capability of the color descriptor - CN (Color Name~\cite{Benavente2008cn}) in describing the clothing colors. The dataset - COLOR335  - is used for evaluation. Examples images of four categories are shown in Fig.~\ref{color:image}. For each category, two images are used for training and the remaining one for testing. One of the most sophisticated color descriptor - `CN' is used to extract color features in the photographs. Since the characteristics of color collocations commonly make clothes more distinctive in visual perception, a regional color co-occurrence framework (RCC) is utilized to describe color collocations. The RCC can be applied to both `CN' features and RGB features, which leads to `CN-RCC' and `RGB-RCC', respectively. In classification, each of the three color descriptors, i.e., `CN', `CN-RCC' and `RGB-RCC', are tested by equipping with 64, 128, and 256 clustering centers, respectively, and the results are shown in Fig.~\ref{color:precision}. We can see that, all the three descriptors achieve classification accuracies higher than 85\%. When the number of clustering centers is tuned to 128 or higher, the accuracies are over 90\%. For `CN', the classification performance consistently improves when the number of clustering centers increases. For `CN-RCC' and `RGB-RCC', the classification accuracies are consistently above 97\% when varying the number of clustering centers from 64 to 256. `CN-RCC' and `RGB-RCC' descriptors have demonstrated their excellence and stability in classifying clothes by encoding the color and color collocations.

\subsection{The capability of texture descriptors in clothing-texture description}
\begin{figure*}[!htbp]
\centering
    \subfigure[]{
        \label{texture:image}
        \includegraphics[width=0.38\linewidth]{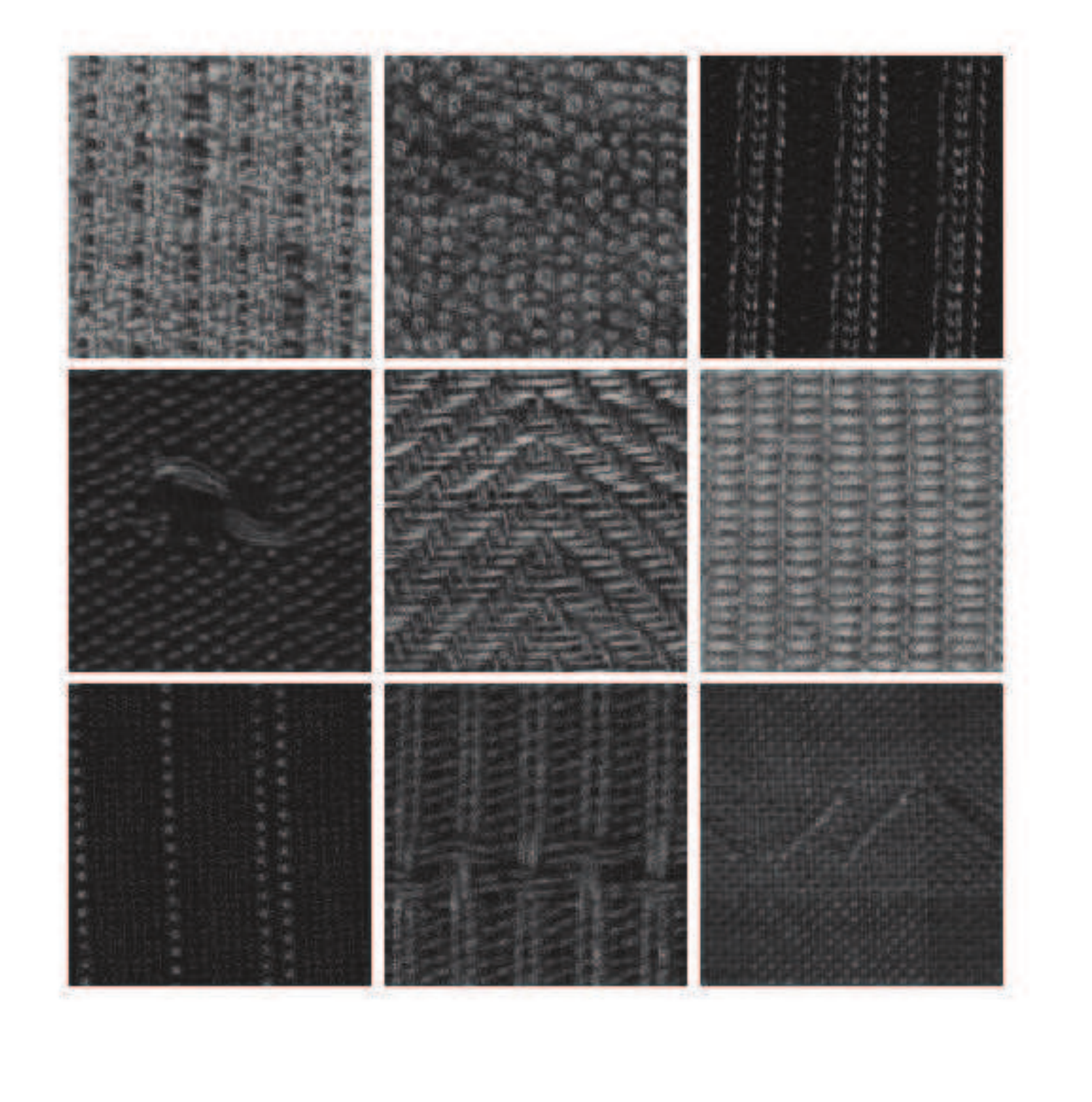}}
    \hspace{0.25in}
    \subfigure[]{
        \label{texture:precision}
        \includegraphics[width=0.47\linewidth]{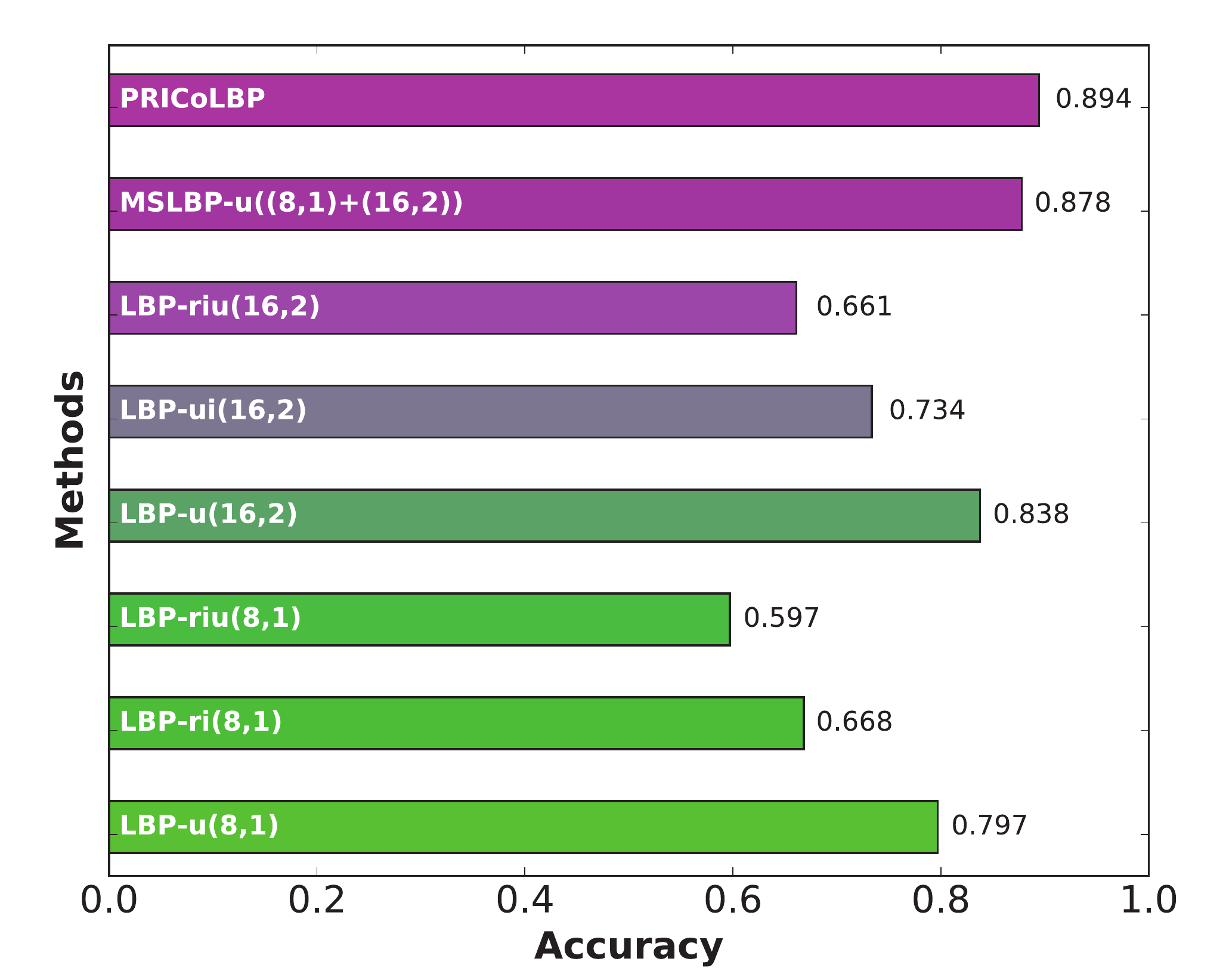}}
    \caption{\textbf{Performance on clothing texture classification.}
    (a) Sample images of nine different texture types from Outex-TC-00021.
    (b) Performances of clothing texture classification using eight different LBP-based texture descriptors.}

\end{figure*}

In Exp.~3, we evaluate the capability of the texture descriptor LBP (Local Binary Patterns~\cite{ojala2002lbp}) in describing the surface texture of clothes. The dataset - Outex-TC-00021 - is used for evaluation. This dataset contains 2,720 images of 68 different texture types, with 40 images per type. Sample images of nine texture types are shown in Fig.~\ref{texture:image}. One half of the images are used for training and the remaining half are used for testing. A total of eight texture descriptors, stemming from LBP, are adopted for performance evaluation, which are LBP$_{u(8,1)}$, LBP$_{ri(8,1)}$, LBP$_{riu(8,1)}$, LBP$_{u(16,2)}$, LBP$_{ri(16,2)}$,  LBP$_{riu(16,2)}$, MSLBP$_{u((8,1)+(16,2))}$\cite{ojala2002lbp}, and PRICoLBP~\cite{qi2014pami}. It can be seen from Fig.~\ref{texture:precision} that, the top three performances are achieved by PRICoLBP, MSLBP$_{u((8,1)+(16,2))}$  and LBP$_{u(16,2)}$, where the accuracies achieved are over 83\%. PRICoLBP outperforms all comparison descriptors and achieves an accuracy of 89.4\%.

\subsection{The influence power of style, color and texture on clothing-fashion updates}
\begin{figure*}[!t]
\centering
    \subfigure[]{
        \label{fstimuli:style}
        \includegraphics[width=0.45\linewidth]{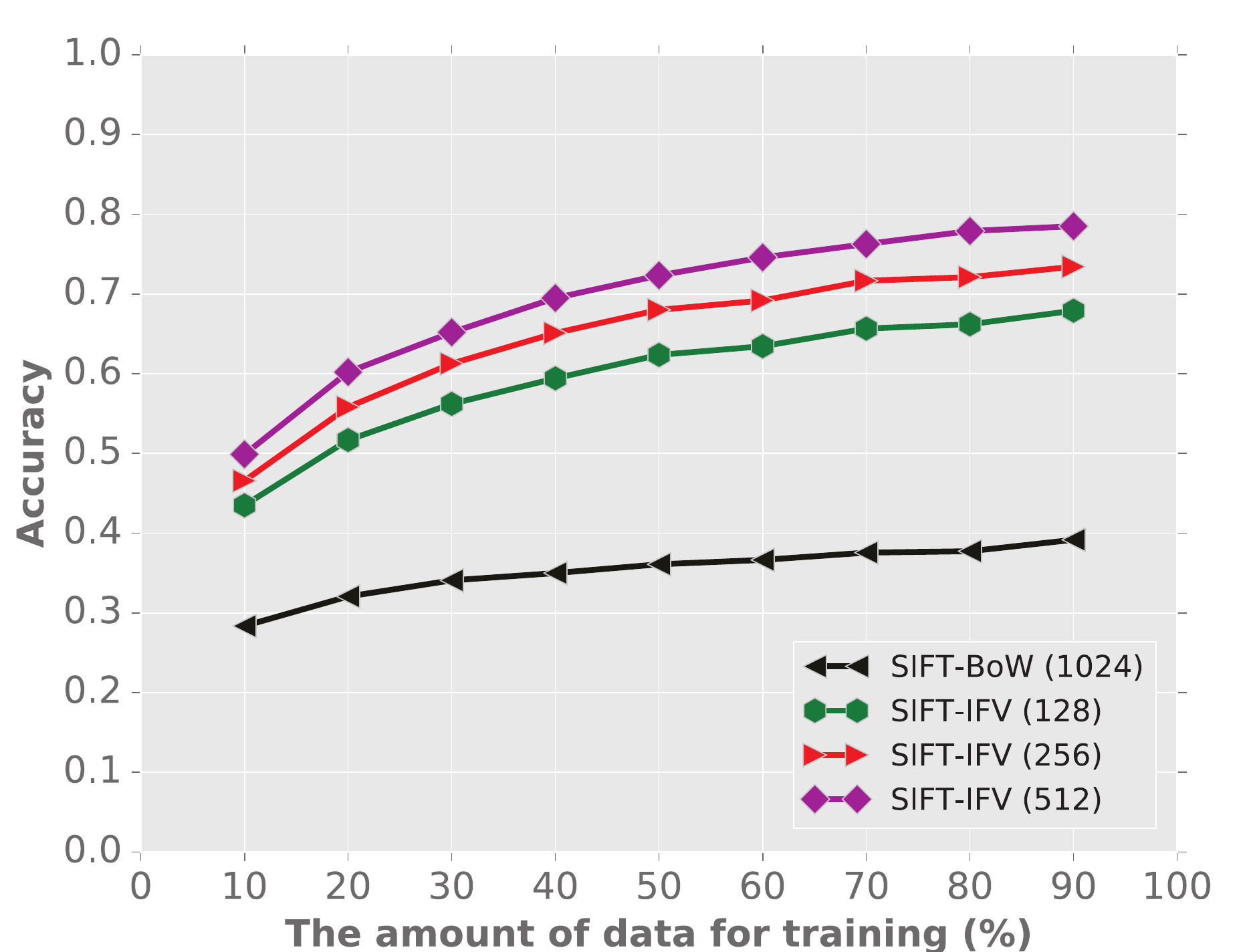}}
    \subfigure[]{
        \label{fstimuli:color}
        \includegraphics[width=0.45\linewidth]{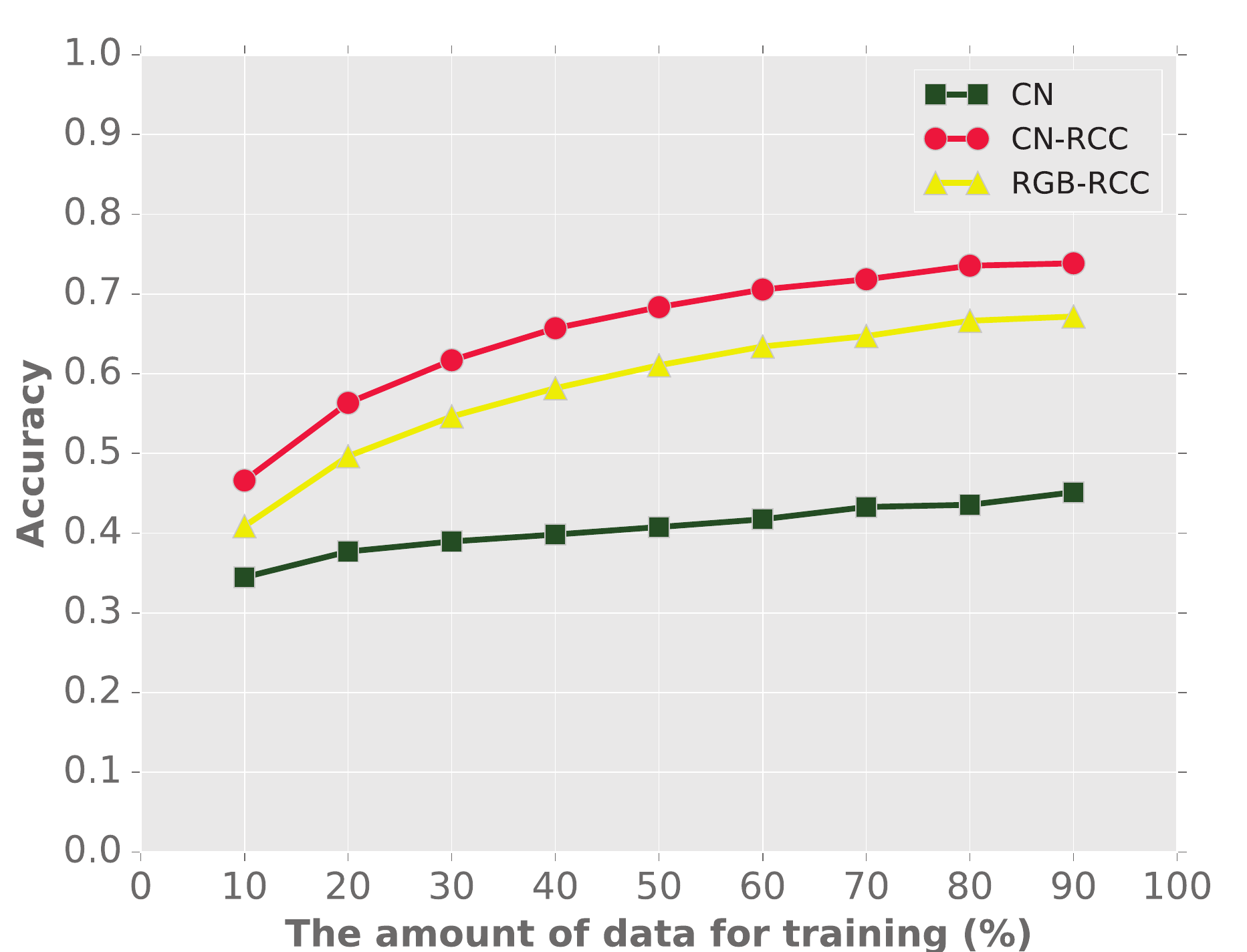}}\\
\vspace{-1mm}
     \subfigure[]{
        \label{fstimuli:texture}
        \includegraphics[width=0.45\linewidth]{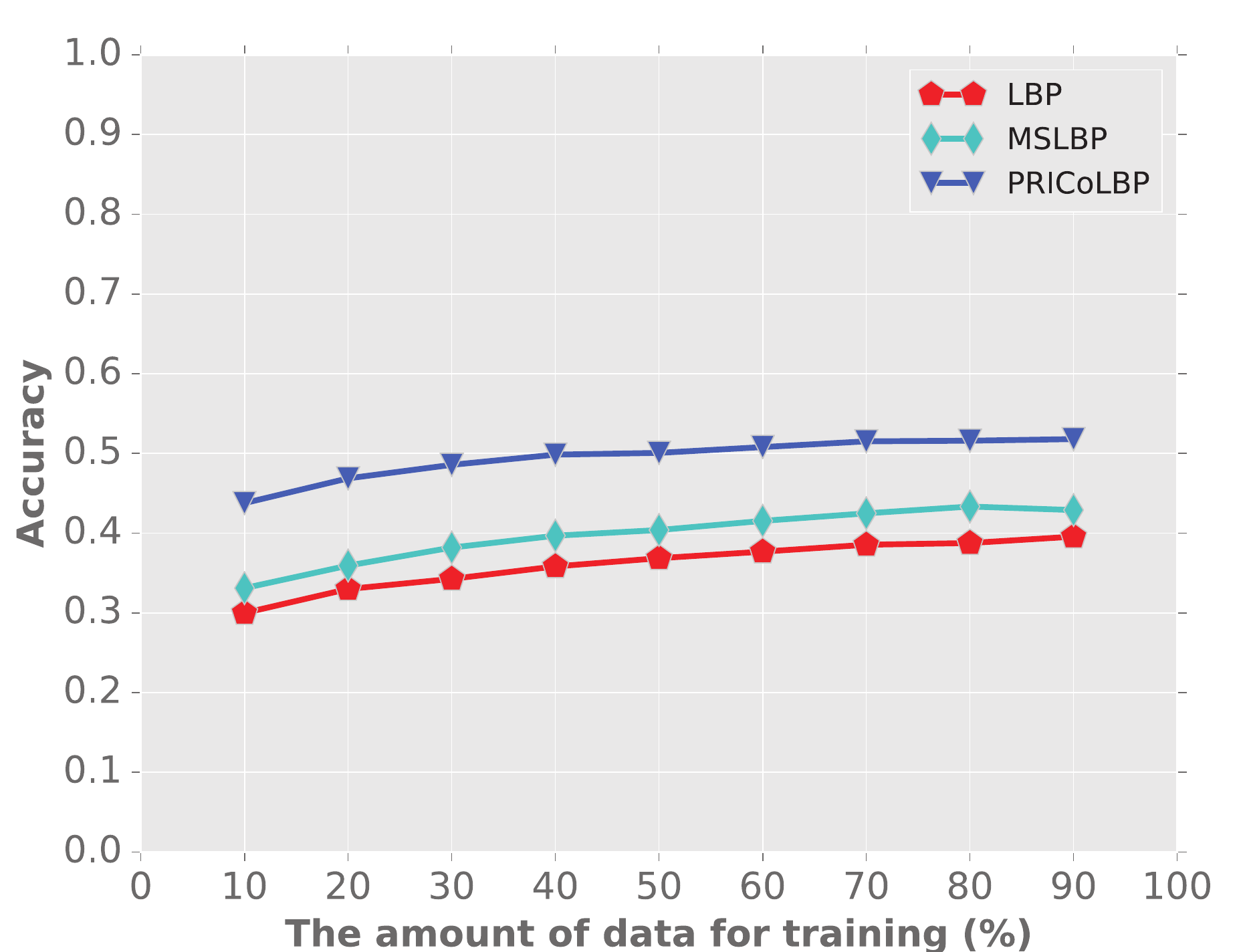}}
    \subfigure[]{
        \label{fstimuli:three}
        \includegraphics[width=0.45\linewidth]{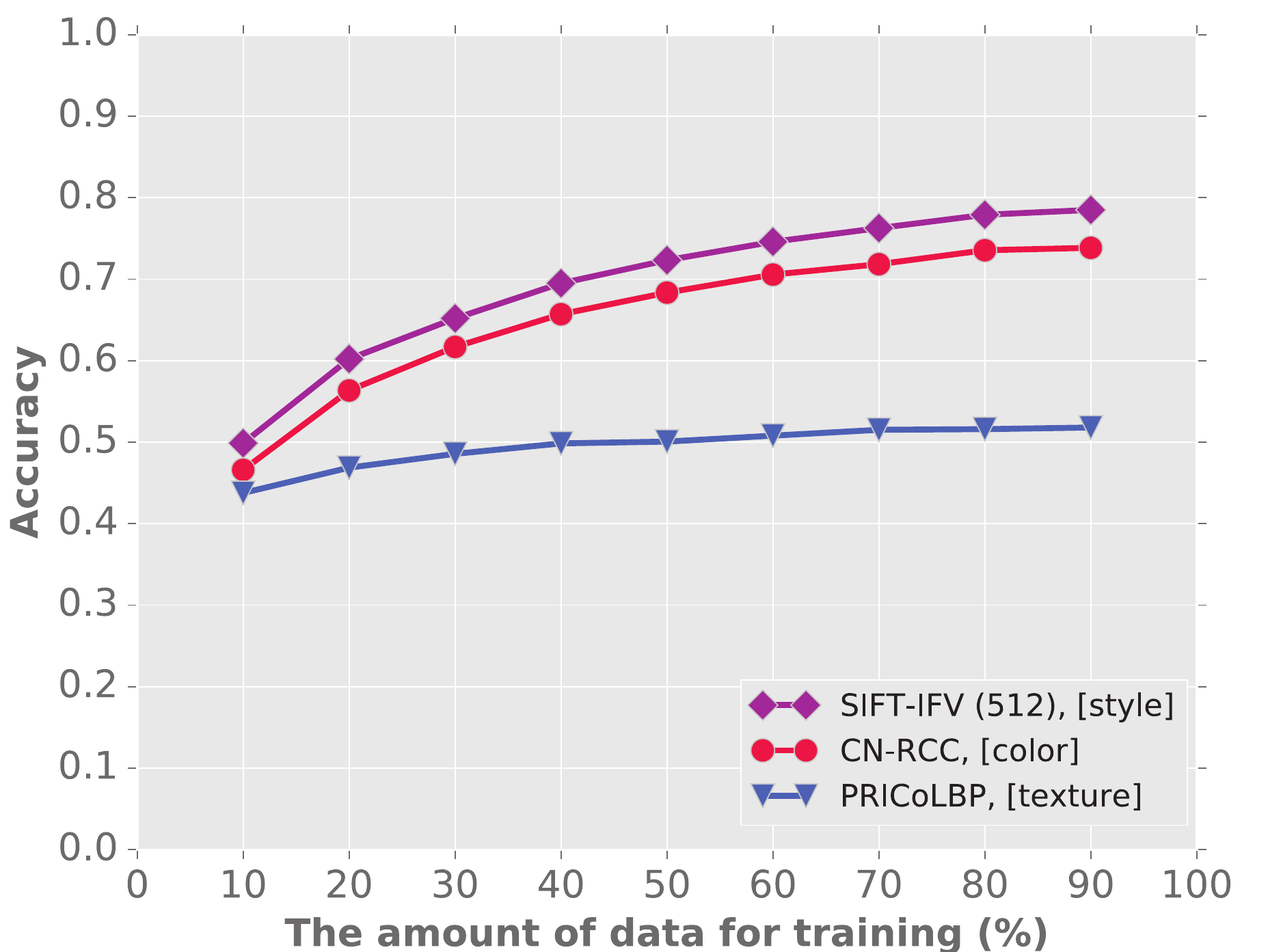}}
\vspace{-1mm}
        \caption{\textbf{Classification performances of color, shape and texture descriptors on FASHION8.}
        (a) Shape descriptors.
        (b) Color descriptors.
        (c) Texture descriptors.
        (d) Comparing of shape, color and texture descriptors, by using the best performed ones in (a), (b) and (c), respectively.}\label{fig:result8}

\end{figure*}

In Exp.~4, the dataset - FASHION8 - is used to analyze the influence power of style, color and texture on clothing-fashion updates. Note that, the 9,339 fashion photographs in FASHION8 have been labeled into 8 categories with respect to the year of the corresponding fashion show. In order to study which visual stimulus, i.e., style, color or texture, places higher/lower influence on the ongoing fashion of clothing, we purposely use each of them to perform the classification. The most capable descriptors derived from Exps. 1, 2, and 3 are selected to quantize the influence of the style, color and texture, respectively. Specifically, we run each algorithm on FASHION8 by varying the amount of training data from 10\% to 90\%, and the results are shown in Fig.~\ref{fig:result8}.

To analyze the influence of style in clothing-fashion updates, four SIFT-based algorithms are employed to extract the style features, which are the `SIFT-BoW (1024)' (using Bag-of-Words with 1024 clustering centers), `SIFT-IFV (128)', `SIFT-IFV (256)', and `SIFT-IFV (512)' (using IFV with 128, 256, and 512 clustering centers, respectively). It can be seen from Fig.~\ref{fstimuli:style} that, `SIFT-IFV (512)' performs the best among the four algorithms. Note that `SIFT-IFV (512)' achieves an accuracy of 72.33\% when using 50\% data for training.

To analyze the influence of color in clothing-fashion updates, three color-based algorithms are employed to extract the color features, which are `CN11', `RGB-RCC' and `CN11-RCC'. All the three descriptors are consistently set with 256 clustering centers in color-codebook construction, and the results are shown in Fig.~\ref{fstimuli:color}. We can see that, `CN11-RCC' is the most representative one that can reflect the power of color on clothing-fashion updates, where an accuracy of 68.36\% is achieved when using 50\% data for training.

To analyze the influence of texture in clothing-fashion updates, three LBP-based algorithms are employed to extract the texture features, which are `LBP$_{u(16,2)}$', `MSLBP$_{u((8,1)+(16,2))}$', and `PRICoLBP'. It can be seen from Fig.~\ref{fstimuli:texture} that, none of these three algorithms achieves an accuracy higher than 60\%. Among the results, the highest accuracy achieved by `PRICoLBP' is 50.07\%, when using 50\% data for training.

For a straightforward comparison, the highest performance achieved by descriptors of style, color and texture, respectively, are plotted in Fig.~\ref{fstimuli:three}, from which we can see that, `SIFT-IFV (512)', which represents the style, stands the highest performance in classification of clothing fashion photographs.  Meanwhile, the descriptor `CN11-RCC' representing the color holds a higher performance than `PRICoLBP', which represents the texture.

\section{Discussion}

In this study, we examine the influence power of style, color and texture on the updating of clothing fashion. It is worth noting that this study is different from existing studies related to clothing and fashions, which are aiming at clothing retrieval~\cite{liu2016deepfashion,hadi2015buy,liu2012street,wang2011clothes,yamaguchi2013paper,Liu2016Fashion}, clothing item parsing or segmentation~\cite{chen2012describing,yamaguchi2012parsing,chen2015deep,kiapour2014hipster,liu2014fashion,yang2014clothing} and clothing-style recognition~\cite{veit2015learning} and popularity recommendation~\cite{yamaguchi2014chic}. We achieve our research goal by ranking the accuracies of fashion classification obtained by the visual stimuli of style, color and texture, respectively. Specifically, the three visual stimuli are represented by the corresponding feature descriptors. Before applying these descriptors to fashion classification on FASHION8, we use three experiments to test and validate the capability of the involved descriptors in describing the corresponding visual stimuli. Exp.~1 shows that the SIFT-based shape descriptors are capable of classifying 130 fashion styles with an accuracy of about 85\%. Exp.~2 shows that the RCC-based color descriptors obtain accuracies over 90\% in classifying 335 clothing categories with different colors. Exp.~3 shows that the fine-tuned LBP-based texture descriptors achieve accuracies over 85\% in classifying 68 surface-texture categories. The high performances demonstrate that these feature descriptors are capable of describing the corresponding visual stimuli possessed by fashion clothing.

Based on the above three experiments and findings, we select the top-three or -four best performed descriptors of shape, color and texture, respectively, to do classification on FASHION8. The findings demonstrate that the stimuli of style, color and texture rank the first, second, and the third in fashion classification, which simply indicates that, in clothing-fashion updates, the style holds a higher influence than the color, and the color holds a higher influence than the texture. Previous studies have suggested that in the primate human visual system, different components of visual information processing, such as shape, color, and texture, are segregated into largely independent parallel pathways~\cite{livingstone1987JON,ishizaki2015scr}. The style, color and texture can independently place some influences on clothing-fashion updates. People may empirically guess that the style and color are more influential than the texture. However they have not yet made any scientific proofs. In our study, assisted by computer vision and machine learning techniques, the quantitative analysis results verified this guess, and in addition have revealed that the style is slightly more influential than the color in clothing-fashion updates.

On the media, color may dominate the analysis and report of the fashion trends, e.g., fashion talks and magazines may keep on talking about fashionable color. However, our study suggest that the highest accuracy obtained by color in fashion classification is about 68\%, which is about 4\% lower that that of style. When using texture descriptors for fashion classification, the highest accuracy obtained is 50.07\%. Since the accuracy is over 50\%, the surface texture also shows an explicit impact on clothing-fashion updates. But comparing with color, the impact of texture is much weaker. We can easily find some real evidences to support this experimental result, e.g., when watching fashion shows or enjoying fashion programs, the first stimulus most people perceive probably be the color rather than the texture. As for fashion magazines and other media materials, texture is always the second or later aspect to be mentioned. Therefore, it is not surprising that the influence power of surface texture is lower than that of color in clothing-fashion updates.

\begin{figure*}[!htbp]
\centering
    \subfigure[]{
        \label{conf_mat:style}
        \includegraphics[width=0.31\linewidth]{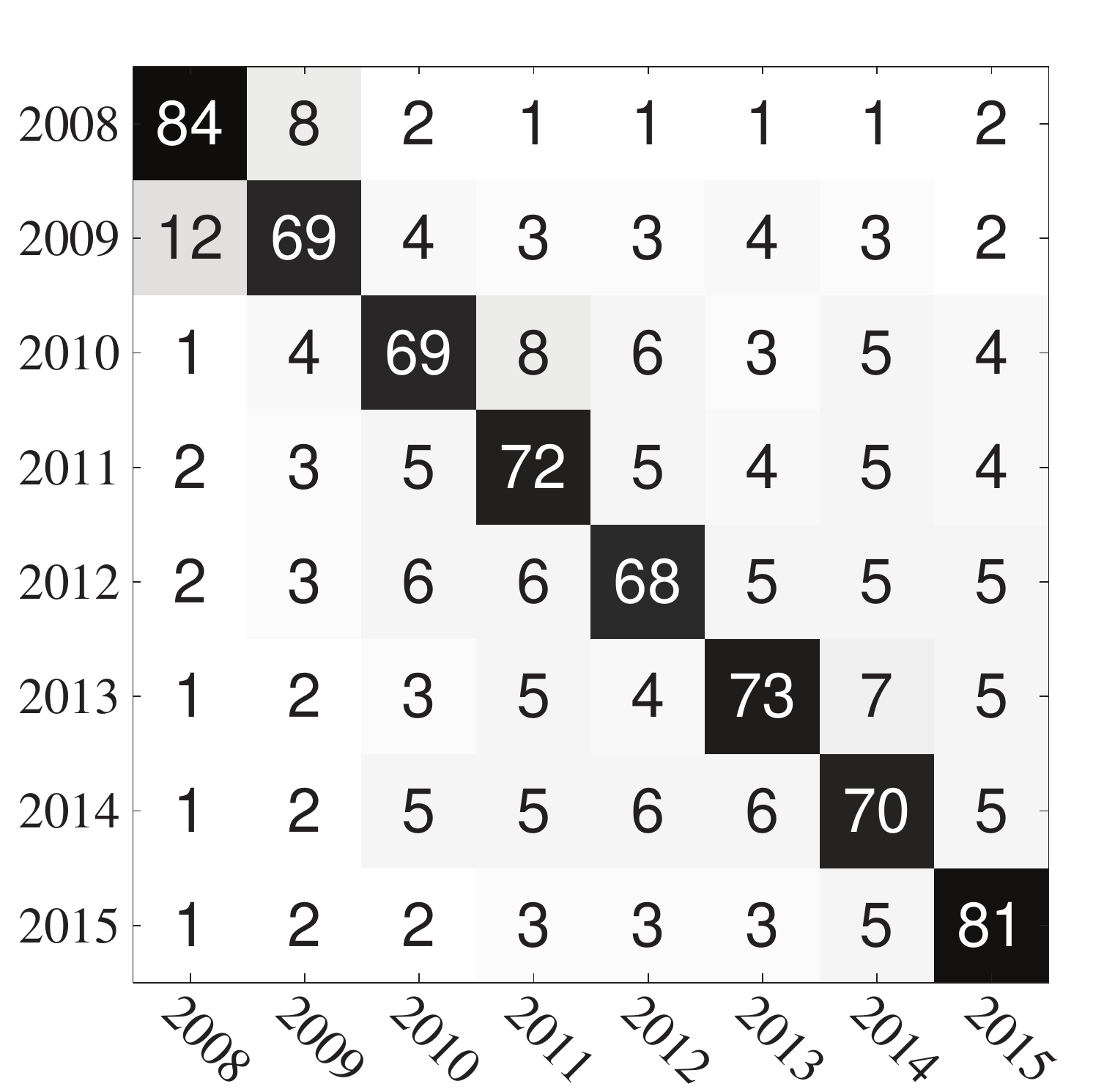}}
    \subfigure[]{
        \label{conf_mat:color}
        \includegraphics[width=0.31\linewidth]{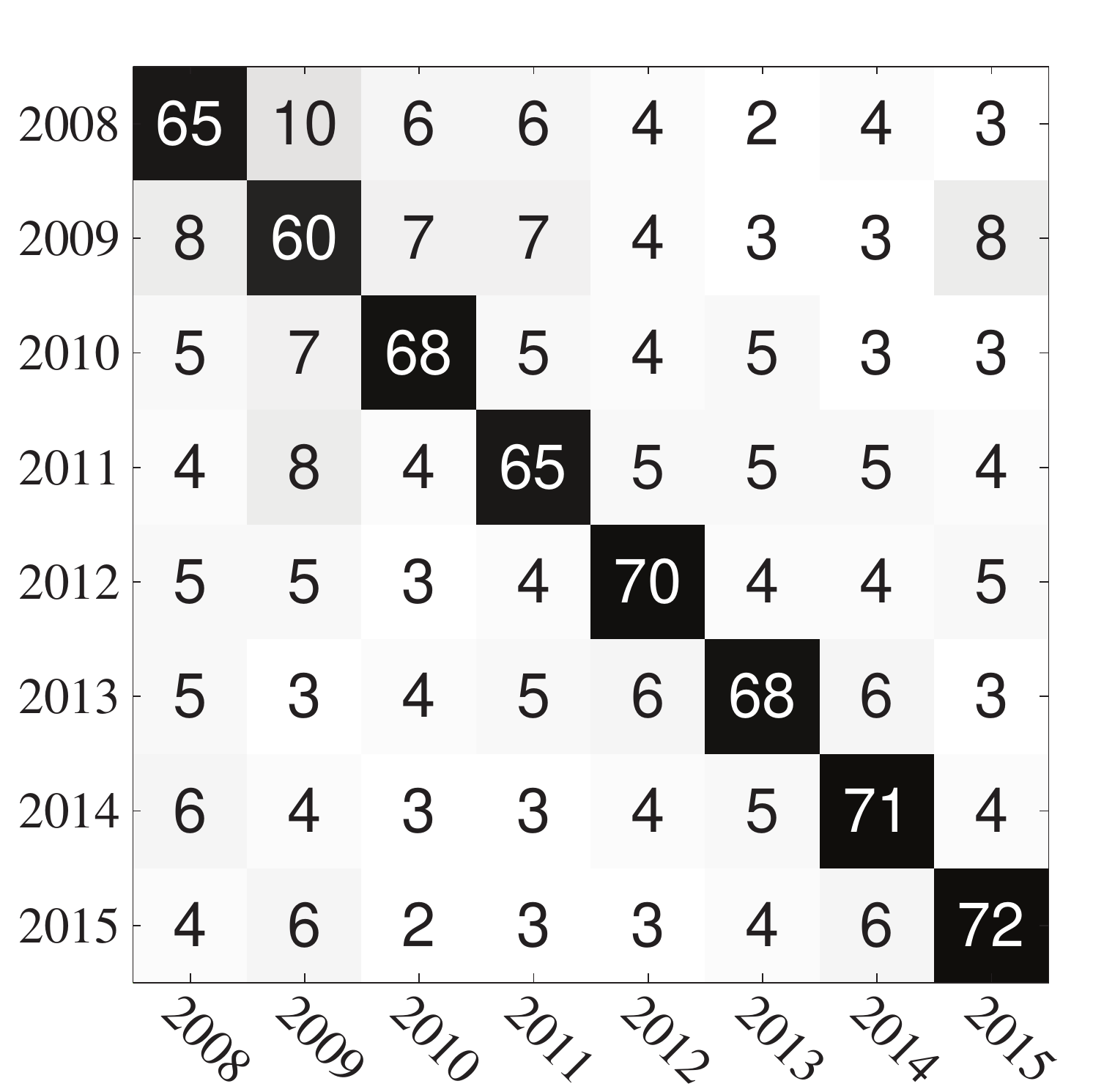}}
     \subfigure[]{
        \label{conf_mat:texture}
        \includegraphics[width=0.31\linewidth]{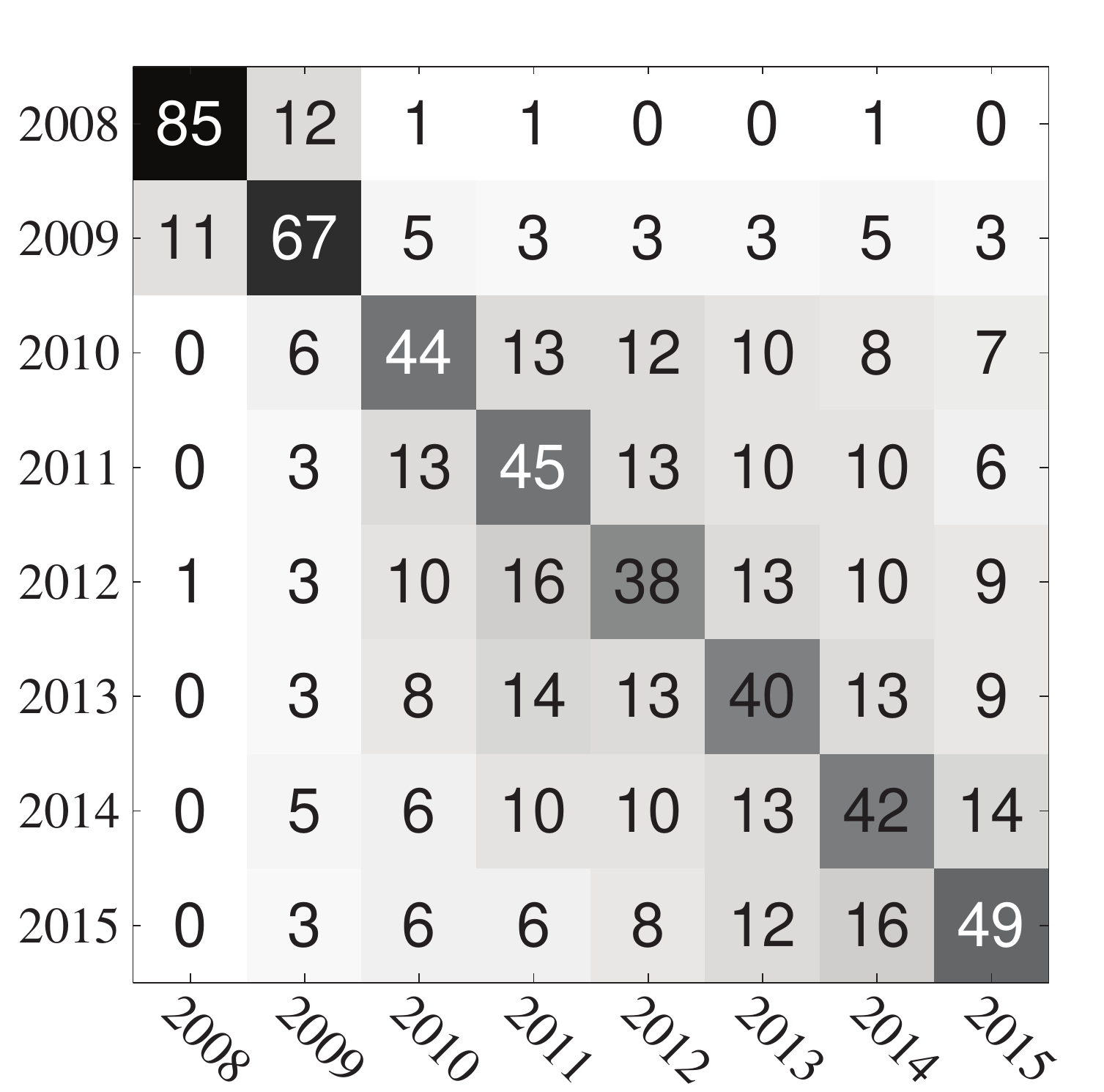}}
\vspace{-1mm}
        \caption{\textbf{Confusion matrices}. The three confusion matrices are constructed based on the classification results by using (a) style, (b) color and (c) texture on dataset FASHION8, respectively.}
\end{figure*}

To further investigate the law of clothing-fashion updates, three confusion matrices are computed for the style-, color- and texture-based fashion classification, as shown in Fig.~\ref{conf_mat:style},~\ref{conf_mat:color}, and~\ref{conf_mat:texture}, respectively.  It can be observed from Fig.~\ref{conf_mat:style} that, in style-based clothing classification, 
most misclassifications occur between adjacent years.
The main reason is that the clothing styles in adjacent years are similar. This indicates that fashion styles do not update in a radical way, i.e., fashion styles evolve continuously year by year. From Fig.~\ref{conf_mat:color}, we can see that, the misclassification is likely to occur between any two years with similar probability. For example, there are 8\% samples in the category 2009 are misclassified into the category 2008, and meanwhile another 8\% are classified into the category 2015. It indicates that the fashion colors do update in a more radical way. From Fig.~\ref{conf_mat:texture}, we can see that, the misclassification rate is higher since 2010, which indicates that, the clothing textures are very similar since 2010. One of the possible reasons is that the clothing materials and manufacture crafts are similar in from 2010 to 2015.

\section{Conclusion}

In this study, computer vision and machine learning techniques were utilized to made a quantitative study to the influence power of style, color and texture on the clothing-fashion updates. First, three experiments were designed to select reliable feature descriptors for clothing style, color and texture description, respectively. Second, the analysis of clothing-fashion updates is mapped into a task of clothing-fashion classification, in which the above selected descriptors were used. Based on the observation that the clothing fashion updates year by year, a ground-truth fashion dataset was constructed by collecting fashion photographs in a continuous eight years. Finally, experiments were conducted to evaluate the capability of three visual stimuli, i.e., style, color and texture, in clothing fashion classification. Experimental results demonstrated that, on clothing-fashion updates, the style held a higher influence than the color, and the color held a higher influence than the texture.

\section*{Acknowledgements}

We would like to thank the members at NIS\&P Lab at Wuhan University for their help in collecting and pre-processing the data used in this work.
The work was supported by the National Natural Science Foundation of China (NSFC) under grant No.~61301277 and the
National Basic Research Program of China under grant No.~2012CB725303.

{\small
\bibliographystyle{ieee}
\bibliography{arxiv_fashion}
}
\flushend

\end{document}